# Seeking multi-thresholds for image segmentation with Learning Automata

**Erik Cuevas, Daniel Zaldivar[1] and Marco Pérez-Cisneros**

Departamento de Ciencias Computacionales
Universidad de Guadalajara, CUCEI
Av. Revolución 1500, Guadalajara, Jal, México
{erik.cuevas, [1]daniel.zaldivar, marco.perez}@cucei.udg.mx

**Abstract**

This paper explores the use of the Learning Automata (LA) algorithm to compute threshold selection for image segmentation as it is a critical preprocessing step for image analysis, pattern recognition and computer vision. LA is a heuristic method which is able to solve complex optimization problems with interesting results in parameter estimation. Despite other techniques commonly seek through the parameter map, LA explores in the probability space providing appropriate convergence properties and robustness. The segmentation task is therefore considered as an optimization problem and the LA is used to generate the image multi-threshold separation. In this approach, one 1-D histogram of a given image is approximated through a Gaussian mixture model whose parameters are calculated using the LA algorithm. Each Gaussian function approximating the histogram represents a pixel class and therefore a threshold point. The method shows fast convergence avoiding the typical sensitivity to initial conditions such as the Expectation-Maximization (EM) algorithm or the complex time-consuming computations commonly found in gradient methods. Experimental results demonstrate the algorithm's ability to perform automatic multi-threshold selection and show interesting advantages as it is compared to other algorithms solving the same task.

***Keywords***: Image segmentation, Learning automata, automatic thresholding, intelligent image processing, Gaussian mixture, Expectation-Maximization, Gradient methods.

## 1. Introduction

Several image processing applications aim to detect and mark relevant features which may be later analyzed to perform several high-level tasks. In particular, image segmentation seeks to group pixels within meaningful regions. Commonly, gray levels belonging to the object, are substantially different from those featuring the background. Thresholding is thus a simple but effective tool to isolate objects of interest; its applications include several classics such as document image analysis, whose goal is to extract printed characters [1,2], logos, graphical content, or musical scores; also it is used for map processing which aims to locate lines, legends, and characters [3]. Moreover, it is employed for scene processing, seeking for object detection, marking [4] and for quality inspection of materials [5,6].

Thresholding selection techniques can be classified into two categories: bi-level and multi-level. In the former, one limit value is chosen to segment an image into two classes: one representing the object and the other one segmenting the background. When distinct objects are depicted within a given scene, multiple threshold values have to be selected for proper segmentation, which is commonly called multilevel thresholding.

A variety of thresholding approaches have been proposed for image segmentation, including conventional methods [7, 8, 9, 10] and intelligent techniques (see for instance [11,12]). Extending the segmentation algorithms to a multilevel approach may cause some inconveniences: (i) they may have no systematic or analytic solution when the number of classes to be detected increases and (ii) they may also show a slow convergence and/or high computational cost [11].

[1] Corresponding author, Tel +52 33 1378 5900, ext. 7715, E-mail: daniel.zaldivar@cucei.udg.mx

In this work, the segmentation algorithm is based on a parametric model holding a probability density function of gray levels which groups a mixture of several Gaussian density functions (Gaussian mixture). Mixtures represent a flexible method of statistical modelling as they are employed in a wide variety of contexts [34]. Gaussian mixture has received considerable attention in the development of segmentation algorithms despite its performance is influenced by the shape of the image histogram and the accuracy of the estimated model parameters [26]. The associated parameters can be calculated considering an approximated maximum aposterior estimation (MAP) or the maximum likelihood (ML) estimation, considering the Expectation Maximization (EM) algorithm [27,31] or Gradient-based methods[30].

The EM algorithm provides a simple alternative procedure for computing posterior density or likelihood functions. However, its slow convergence speed has been pointed out as the most serious practical problem [28]. When the EM is used aside Gaussian mixtures, the convergence speed depends on the separation of component populations within such mixture. Therefore, the EM algorithms are very sensitive to the choice of the initial values [29]. Moreover, the EM algorithm also tends to converge to a local minima [35,36]. A feasible way for solving this problem is to choose several sets of initial values, applying the EM algorithm and finally choosing the best outcome-set as the best estimation. By doing so, it can alleviate the influence of the initial values on the algorithm but increasing the computational cost. Additionally, the EM algorithm fails to converge if one or some variances of the Gaussian mixture approach to zero as it has been demonstrated when big objects with uniform intensities have undergone segmentation [26].

On the other hand, Gradient-based methods are also computationally expensive and may easily get stuck within local minima [26]. Redner and Walkerin argued in [28] -a widely-cited article, that Newtonian methods (such as Levenberg-Marquardt, LM) should generally be preferred over EM particularly for unconstrained optimization problems. However, they must be modified in order to be used within Gaussian mixtures, where there are probabilistic constraints on the parameters [32] that may result in singularities. In the parameter space of mixture models, the singularities occur when two or more components are exactly overlapped and they can be dismissed as a single component. Recent works in [29] and [33] have shown that singularities cause slow convergence in Newtonian and quasi-Newtonian methods while they are applied to determine parameters of Gaussian mixtures.

Despite gradient-based methods and the EM algorithm seem to have different mechanisms for parameter updating. Xu and Jordan [32] have established a relationship between the gradient of the log likelihood and the updating step within the parameter space while using the EM algorithm. They found that the EM algorithm can be viewed as a variable metric gradient algorithm with a projection matrix changing at each step and behaving just like a function of the current parameter value. In the EM algorithm, the new value of the parameter $k+1$ is thus chosen close to the previous value $k$ mimicking gradients methods. Therefore, the updating rule may get the EM algorithm easily stuck within local minima [33].

In this paper, an alternative approach using an optimization algorithm based on learning automata for determining the parameters of a Gaussian mixture is presented. The Learning Automata (LA) [24,25] is an adaptive decision making method that operate in unknown random environments while progressively improve their performance via a learning process. Since LA theorists study the optimization under uncertainty, it is very useful for optimization of multi-modal functions when the function is unknown and only noise-corrupted evaluations are available. In such algorithms, a probability density function, which is defined over the parameter space, is used to select the next point. The reinforcement signal (objective function) and the learning algorithm are used by the learning automata (LA) to update the probability density function at each stage. LA has been successfully applied to solve different sorts of engineering problems such as pattern recognition [17], adaptive control [18] signal processing [19] and power systems [20].

One main advantage of the LA method is that it does not need knowledge of the environment or any other analytical reference to the function to be optimized. Additionally, one interesting advantage of LA lies on the fact that it offers fast convergence mainly when it is considered for the estimation of many parameters [21]. Other methods such as Gradient and the EM which make use of iterative updating procedures within the parameter space, may exhibit slow convergence or local minima trapping. However LA is focused on the probability space [22] leading to global optimization [23] by allowing any element of the action set (or parameter) to be chosen. This fact actually makes LA insensitive to initial values.

Recently, more effective LA-based algorithms have been proposed for multimodal complex function optimization [19, 21, 22, 23]. It has also been experimentally shown that the performance of such optimization algorithms is comparable to or better than the genetic algorithm (GA) in [22]. On the other hand, the algorithm known as *continuous action reinforcement learning automata* (CARLA) [37], has been used for parameter identification of particularly complex systems, showing the effectiveness of the approach with interesting results on adaptive control [37,38,39,40] and digital filter design [19].

In this paper, the segmentation process is considered as an optimization problem approximating the 1-D histogram of a given image by means of a Gaussian mixture model. The operation parameters are calculated through the CARLA algorithm. Each Gaussian contained within the histogram represents a pixel class and therefore belongs to the thresholding points. The experimental results, presented in this work, demonstrate that LA exhibits fast convergence, relative low computational cost and no sensitivity to initial conditions by keeping an acceptable segmentation of the image, i.e. a better mixture approximation in comparison to the EM or gradient based algorithms.

The paper is organized as follows. Section 2 presents the Gaussian approximation to the histogram while Section 3 introduces the LA algorithm. Section 4 shows the most important implementation issues. Experimental results for the proposed approach are presented in Section 5 and some relevant conclusions are discussed in Section 6.

## 2. Gaussian approximation

Let consider an image holding $L$ gray levels $[0,\ldots,L-1]$ whose distribution is displayed within a histogram $h(g)$. In order to simplify the description, the histogram is normalized just as a probability distribution function, yielding:

$$h(g)=\frac{n_g}{N},\quad h(g)>0,$$
$$N=\sum_{g=0}^{L-1} n_g,\ \text{and} \sum_{g=0}^{L-1} h(g)=1, \tag{1}$$

where $n_g$ denotes the number of pixels with gray level $g$ and $N$ being the total number of pixels in the image. The histogram function can thus be contained into a mix of Gaussian probability functions of the form:

$$p(x)=\sum_{i=1}^{K} P_i \cdot p_i(x)=\sum_{i=1}^{K}\frac{P_i}{\sqrt{2\pi\sigma_i}}\exp\left[\frac{-(x-\mu_i)^2}{2\sigma_i^2}\right] \tag{2}$$

with $P_i$ being the probability of class $i$, $p_i(x)$ being the probability distribution function of gray-level random variable $x$ in class $i$, with $\mu_i$ and $\sigma_i$ being the mean and standard deviation of the $i$-th probability distribution function and $K$ being the number of classes within the image. In addition, the constraint $\sum_{i=1}^{K} P_i = 1$ must be satisfied.

The mean square error is used to estimate the $3K$ parameters $P_i$, $\mu_i$ and $\sigma_i$, $i$ = 1, . . ,$K$. For instance, the mean square error between the Gaussian mixture $p(x_i)$ and the experimental histogram function $h(x_i)$ is now defined as follows:

$$J = \frac{1}{n}\sum_{j=1}^{n}\left[p(x_j) - h(x_j)\right]^2 + \omega \cdot \left|\left(\sum_{i=1}^{K} P_i\right) - 1\right| \tag{3}$$

Assuming an $n$-point histogram as in [13] and $\omega$ being the penalty associated with the constrain $\sum_{i=1}^{K} P_i = 1$. In general, the estimation of the parameters that minimize the square error produced by the Gaussian mixture is not a simple problem. A straightforward method is to consider the partial derivatives of the error function to zero, obtaining a set of simultaneous transcendental equations [13]. However, an analytical solution is not available considering the non-linear nature of the equations. The algorithms therefore make use of an iterative approach which is based on the gradient information or maximum likelihood estimation, just like the EM algorithm. Unfortunately, such methods may also get easily stuck within local minima.

For the EM algorithm and the gradient-based methods, the new parameter point lies within a neighbourhood distance of the previous parameter point. However, this is not the case for the LA's adaptation algorithm which is based on stochastic principles. The new operating point is thus determined by a parameter probability function and therefore it can be far from the previous operating point. This gives the algorithm a higher ability to locate and pursue a global minimum.

It has been shown by many papers in the literature that intelligent approaches may actually provide a satisfactory performance for image processing problems [11, 12, 14, 15, 16]. The LA approach was chosen aiming into find appropriate parameters and their corresponding threshold values, yet relying on the LA convergence characteristics and its immunity to initial values.

## 3. LEARNING AUTOMATA (LA)

LA operates by selecting actions via a stochastic process. Such actions operate within an environment while being assessed according to a measure of the system performance. Figure 1a shows the typical learning system architecture. The automaton selects an action (**X**) probabilistically. Such actions are applied to the environment and the performance evaluation function provides a reinforcement signal $\beta$. This is used to update the automaton's internal probability distribution whereby actions that achieve desirable performance are reinforced via an increased probability. Likewise, those underperforming actions are penalised or left unchanged depending on the particular learning rule which has been employed. Over time, the average performance of the system will improve until a given limit is reached. In terms of optimization problems, the action with the highest probability would correspond to the global minimum as demonstrated by rigorous proofs of convergence available in [24] and [25].

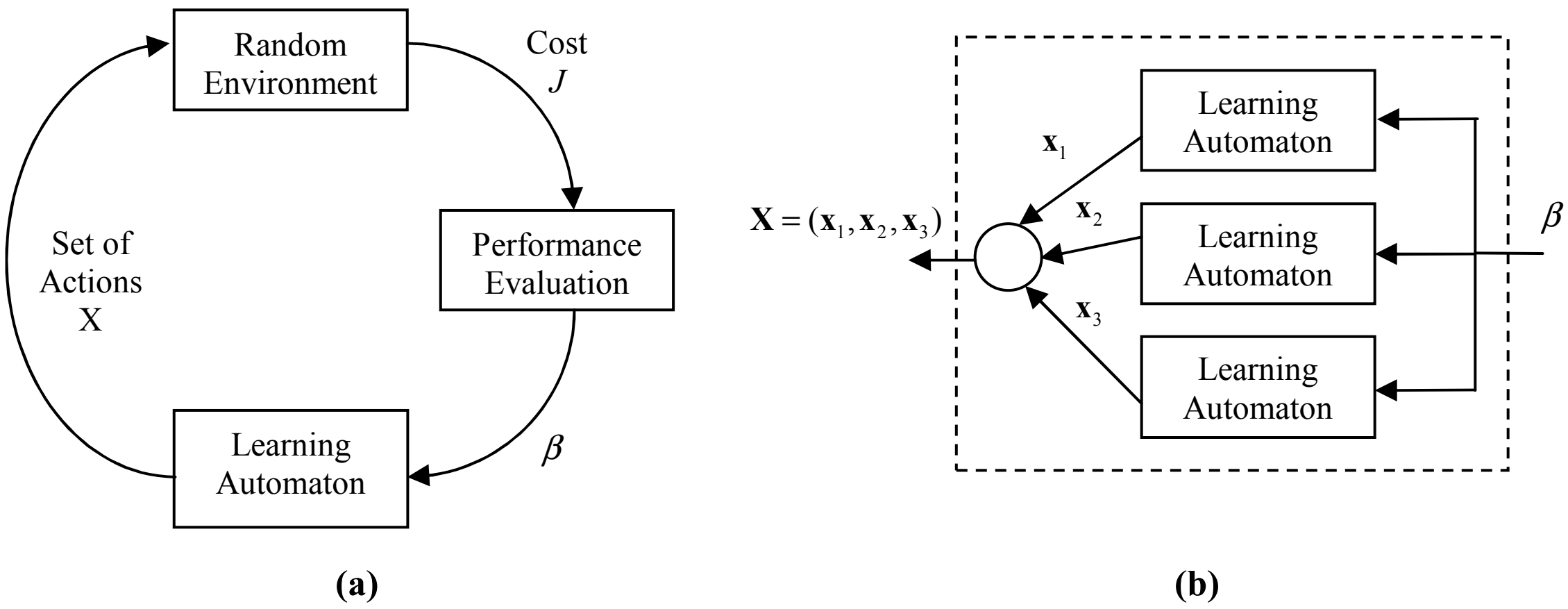

**Figure 1.** (a) Reinforcement learning system and (b) Interconnected automata.

A wide variety of learning rules have been reported in the literature. One of the most widely used algorithms is the linear reward/inaction ( $L_{RI}$ ) scheme, which has been shown to guaranteed convergence properties (see [24,25]). In response to action $\mathbf{x}_i$ , which is selected at time step *k*, the probabilities are updated as follows:

$$p_i(n+1) = p_i(n) + \theta \cdot \beta(n) \cdot (1 - p_i(n))$$
$$p_j(n+1) = p_j(n) - \theta \cdot \beta(n) \cdot p_j(n)\,, \text{ if } i \neq j \tag{4}$$

being $\theta$ a learning rate parameter and $0 < \theta < 1$ and $\beta \in [0,1]$ the reinforcement signal; $\beta = 1$ indicates the maximum reward and $\beta = 0$ is a null reward. Eventually, the probability of successful actions will increase to become close to unity. In case that a single and foremost successful action prevails, the automaton is deemed to have converged.

With a large number of discrete actions, the probability of selecting any particular action becomes low and the convergence time can become excessive. To avoid this, LA can be connected in a parallel setup as the one shown in Figure 1b. Each automaton operates a smaller number of actions and the 'team' works together in a co-operative manner. This scheme can also be used where multiple actions are required.

Discrete stochastic LA can be used to determine global optimal parameters for optimization applications within multi-modal mean-square error surfaces. However, the discrete nature of the automata requires the discretization of a continuous parameter space while the quantization level tends to reduce the convergence rate. Therefore, a sequential approach is adopted for the CARLA implementation [37], overcoming the problem by means of an initial coarse quantization. The method may be refined again by using a re-quantization around the most successful action later on.

### 3.1 CARLA Algorithm

The continuous action reinforcement learning automata (CARLA) is developed as an extension of the discrete stochastic LA for applications involving searching of continuous action space in a random environment [19]. Several CARLA can be connected in parallel similarly to discrete automata (Figure 1b), in order to search multidimensional action spaces. Although the interconnection of the automata is through the environment, no direct inter-automata communication exists. The automaton's discrete probability distribution is replaced by a continuous probability density function which is used as the basis for action selection. It operates a reward/inaction learning rule similar to the discrete LA shown in Equation 4. Successful actions receive an increase on the probability of being selected in the future via a Gaussian neighborhood function which augments the probability density in the vicinity of such successful action. Table 1 shows the generic pseudo-code for the CARLA algorithm. The initial probability distribution can be chosen as being uniform over a desired range. After a considerable number of iterations, it converges to a probability distribution with a global maximum around the best action value [23].

| **CARLA Algorithm** |
| --- |
| Initialize the probability density function to a uniform distribution |
| Repeat |
| Select an action using its probability density function |
| Execute action on the environment |
| Receive cost/reward for previous action |
| Update performance evaluation function $\beta$ |
| Update probability density function |
| Until stopping condition is reached. |

**Table 1.** Generic pseudo-code for the CARLA algorithm.

If action $x$ (parameter) is defined over the range $(x_{\min}, x_{\max})$, the probability density function $f(x,n)$ at iteration $n$ is updated according to the following rule:

$$f(x,n+1) = \begin{cases} \alpha \cdot [f(x,n) + \beta(n) \cdot H(x,r)] & \text{if } x \in (x_{\min}, x_{\min}) \\ 0 & \text{otherwise} \end{cases} \tag{5}$$

With $\alpha$ being chosen to re-normalize the distribution according to the following condition

$$\int_{x_{\min}}^{x_{\max}} f(x,n+1)dx = 1 \tag{6}$$

with $\beta(n)$ being again the reinforcement signal from the performance evaluation and $H(x,r)$ being a symmetric Gaussian neighborhood function centered on $r = x(n)$. It yields

$$H(x,r) = \lambda \cdot \exp\left(-\frac{(x-r)^2}{2\sigma^2}\right) \tag{7}$$

with $\lambda$ and $\sigma$ being parameters that determine the height and width of the neighborhood function. They are defined in terms of the range of actions as follows:

$$\sigma = g_w \cdot (x_{\max} - x_{\min}) \tag{8}$$

$$\lambda = \frac{g_h}{(x_{\max} - x_{\min})} \tag{9}$$

The speed and resolution of learning are thus controlled by free parameters $g_w$ and $g_h$. Let action $x(n)$ be applied to the environment at iteration $n$, returning a cost or performance index $J(n)$. Current and previous costs are stored as a reference set $R(n)$. The median and minimum values $J_{\text{med}}$ and $J_{\min}$ may thus be calculated by means of $\beta(n)$, which is defined as follows:

$$\beta(n) = \max\left\{0, \frac{J_{\text{med}} - J(n)}{J_{\text{med}} - J_{\min}}\right\} \tag{10}$$

To avoid problems with infinite storage requirements and to allow the system to adapt to changing environments, only the last $m$ values of the cost functions are stored in $R(n)$. Equation (10) limits $\beta(n)$ to values between 0 and 1 and only returns nonzero values for those costs that are below the median value. It is easy to understand how $\beta(n)$ affects the learning process as follows: during the learning, the performance and the number of selecting actions can be wildly variable, generating extremely high computing costs. However, $\beta(n)$ is insensitive to such extremes and to high values of $J(n)$ resulting from a poor choice of actions. As the learning continues, the automaton converges towards more worthy regions of the parameter space as such actions are chosen to be evaluated more often. When more of such responses are being received, $J_{\text{med}}$ gets reduced. Decreasing $J_{\text{med}}$ in $\beta(n)$ effectively enables the automaton to refine its reference around better responses (previously received), and hence resulting in a better discrimination between selected actions.

In order to define an action value $x(n)$ which has been associated to a given probability density function, an uniformly distributed pseudo-random number $z(n)$ is generated within the range of [0,1]. Simple interpolation is thus employed to equate this value to the cumulative distribution function:

$$\int_{x_{\min}}^{x(n)} f(x,n)dx = z(n) \tag{11}$$

For implementation purposes, the distribution is stored at discrete points with an equal inter-sample probability. Linear interpolation is used to determine values at intermediate positions (see full details in [19]).

## 4. Implementation

Four different pixel classes are used to segment the images. The idea is to show the effectiveness of the algorithm and its performance against other algorithms solving the same task. The implementation can easily be transferred to cases with a greater number of pixel classes.

To approach the histogram of an image by 4 Gaussian functions (one for each pixel class), it is necessary to calculate the optimum values of the 3 parameters ($P_i$, $\mu_i$ and $\sigma_i$) for each Gaussian function (in this case, 12 values according to equation 2). This problem can be solved by optimizing equation 3, considering that function $p(x)$ gathers 4 Gaussian functions.

The parameters to be optimized are summarized in Table 2., with $k_P^i$ being the parameter representing the a priori probability (*P*), $k_\sigma^i$ holding the variance ($\sigma$) and $k_\mu^i$ representing the expected value ($\mu$) of the Gaussian function *i.*

| Parameters | | | Gaussian |
|---|---|---|---|
| $k_P^1$ | $k_\sigma^1$ | $k_\mu^1$ | 1 |
| $k_P^2$ | $k_\sigma^2$ | $k_\mu^2$ | 2 |
| $k_P^3$ | $k_\sigma^3$ | $k_\mu^3$ | 3 |
| $k_P^4$ | $k_\sigma^4$ | $k_\mu^4$ | 4 |

**Table 2.** Parameters to be optimized by the LA algorithm.

In the LA optimization, each parameter is considered like an Automaton which is able to choose actions. Such actions correspond to values assigned to the parameters by a probability distribution within the interval. All intervals considered in this work are defined as $k_P^i \in [0,0.5]$, $k_\sigma^i \in [0,128]$, and $k_\mu^i \in [0,255]$.

For this 12-dimensional problem, 12 different automatons will be created to represent parametric approach of the corresponding histogram. One of the main advantages of the LA algorithm regarding multi-dimensional problems is that the automatons are coupled only through the environment, thus each automaton operates independently during optimization.

Thus, at each instant *n,* each automaton chooses an action according to their probability distribution which can be represented in a vector $A(n)$={ $k_P^1$, $k_\sigma^1$, $k_\mu^1$ …, $k_P^4$, $k_\sigma^4$, $k_\mu^4$ }. This vector represents a certain approach to the histogram. Then, the quality of the approach is evaluated (according to Equation 3) and converted into a reinforcement signal $\beta(n)$ (through Equation 10). After the reinforcement value $\beta(n)$ is defined as a product of the elected approach *A(n),* the distribution of probability is updated for $n$+1 of each automaton (according to the Equation 5). To simplify parameters in Equation 8 and 9, they will take the same value for the 12 automatons, such that $g_w = 0.02$ and $g_h$ =0.3. In this work, the optimization process considers a limit up to 2000 iterations.

The optimization algorithm can thus be described as follows:

**i** Set iteration number *n*=0.

**ii** Define the action set $A(n)$={ $k_P^1$, $k_\sigma^1$, $k_\mu^1$ …, $k_P^4$, $k_\sigma^4$, $k_\mu^4$ } such that $k_P^i \in [0,0.5]$, $k_\sigma^i \in [0,128]$ and $k_\mu^i \in [0,255]$.

**iii** Define probability density functions at iteration *n*: $f(k_P^i, n)$, $f(k_\sigma^i, n)$ and $f(k_\mu^i, n)$

**iv** Initialize $f(k_P^i, n)$, $f(k_\sigma^i, n)$ and $f(k_\mu^i, n)$ as a uniform distribution between the defined limits.

**v** Repeat while $n \leq 2000$

- **(a)** Using a pseudo-random number generator for each automaton, select $z_P^i(n)$, $z_\sigma^i(n)$ and $z_\mu^i(n)$ uniformly between 0 and 1.
- **(b)** Select $k_P^i \in [0,0.5]$, $k_\sigma^i \in [0,128]$ and $k_\mu^i \in [0,255]$ where the area under the probability density function is $\int_0^{k_P^i(n)} f(k_P^i, n) = z_P^i(n)$, $\int_0^{k_\sigma^i(n)} f(k_\sigma^i, n) = z_\sigma^i(n)$ and $\int_0^{k_\mu^i(n)} f(k_\mu^i, n) = z_\mu^i(n)$.
- **(c)** Evaluate the performance using Eq. (3).
- **(d)** Obtain the minimum, $J_{\min}$, and median, $J_{\text{med}}$ of $J(n)$.
- **(e)** Evaluate $\beta(n)$ via Eq. (10).
- **(f)** Update the probability density functions $f(k_P^i, n)$, $f(k_\sigma^i, n)$ and $f(k_\mu^i, n)$ using Eq. (5).
- **(g)** Increment iteration number *n*.

The learning system searches within the 12-dimensional parameter space aiming for reducing the values of $J$ in Equation 3.

The final step is to determine the optimal threshold values $T_i$. In this case, the pixel classification corresponds to the maximum likelihood (ML) estimator. The classes can be determined by simple thresholding following standard methods, just as it is illustrated in the Figure 2.

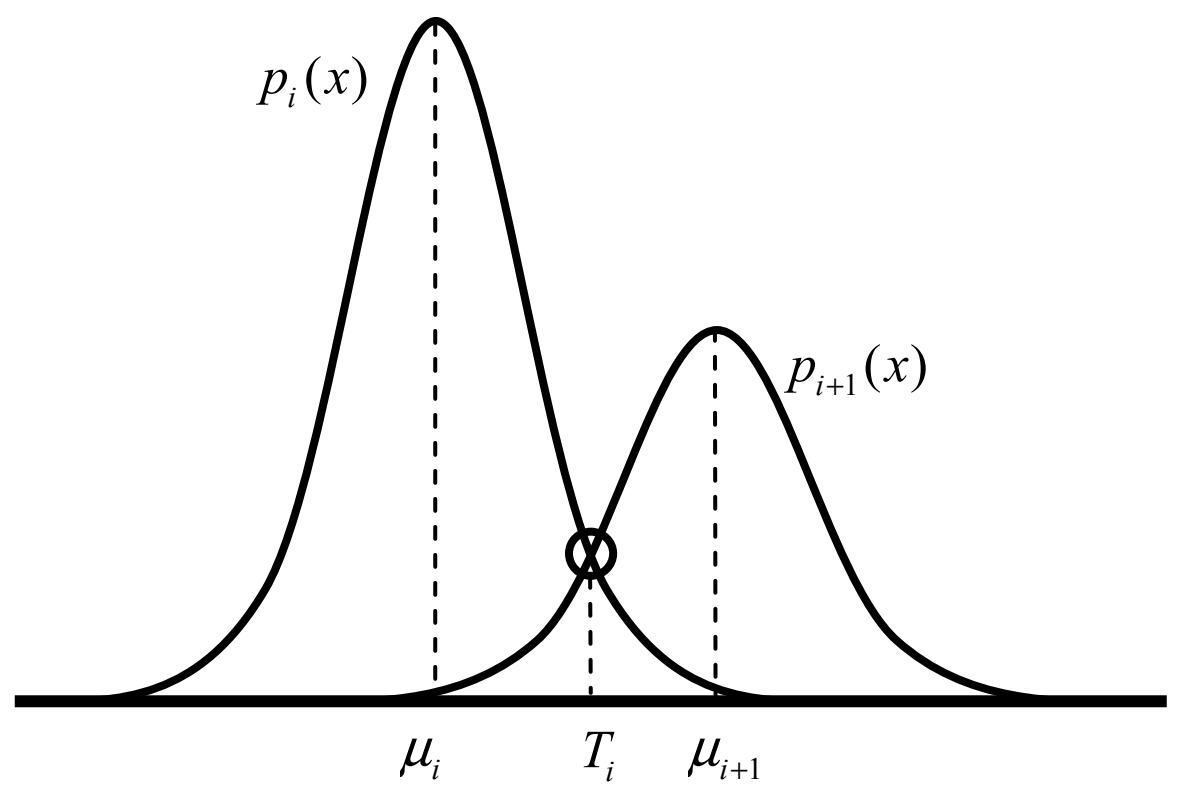

**Figure 2.** Thresholding points determination.

## 5. Experimental results

This section presents the experimental work with the LA algorithm. The discussion is divided into two parts: the first one shows the performance of the proposed LA algorithm while the second discusses on a comparison between the LA segmentator, the EM algorithm and the Levenberg-Marquardt method.

### *5.1 LA algorithm performance in image segmentation*

This section presents two experiments to analyze the LA's performance considering a segmentation mixture of four classes while the original histogram of the image is approached by the LA method. In order to test consistency, 10 independent repetitions are made for each experiment.

The first test considers the histogram shown by Figure 3b while Figure 3a presents the original image. After applying the LA algorithm (as it is explained in the previous section), a minimum is obtained (Equation 3), as the point is defined by $k_P^1$=0.094, $k_\sigma^1$=6, $k_\mu^1$=15, $k_P^2$=0.1816, $k_\sigma^2$=29, $k_\mu^2$=63, $k_P^3$=0.2733, $k_\sigma^3$=10, $k_\mu^3$=93, $k_P^4$=0.4503, $k_\sigma^4$=30, and $k_\mu^4$=163. The values of such parameters define four different Gaussian functions which are clearly visible in Figure 4. The original histogram and its approximation by the Gaussian mixture are visually compared in Figure 5.

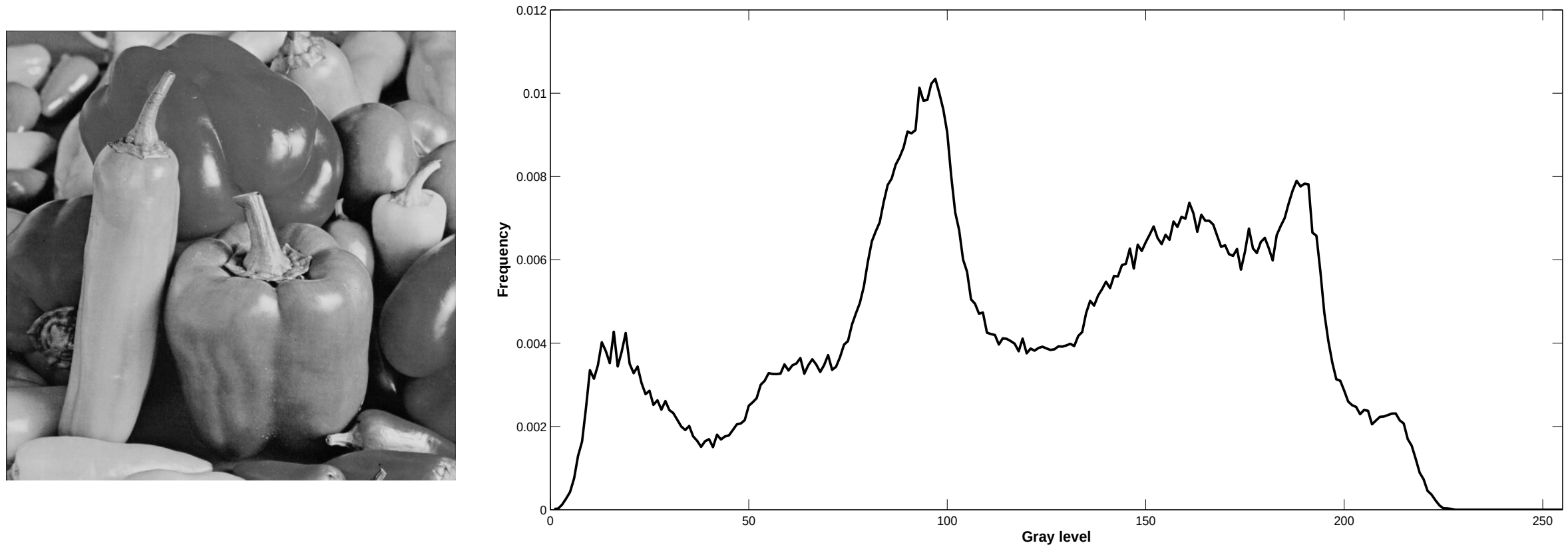

(a) (b)

**Figure 3.** (a) Original image used on the first experiment, (b) and its histogram.

The evolution of the probability density parameters which in turn represent the expected values $f(k_\mu^1, n)$, $f(k_\mu^2, n)$, $f(k_\mu^3, n)$ and $f(k_\mu^4, n)$ of the Gaussian functions are shown in Figure 6. It can be seen that most of the convergence is achieved at the first 1050 iterations, as subsequent steps yield a bit of sharpening in the distribution's shape. The final highest probability value obtained from the distribution (*n*=2000) corresponds to the final parameter value.

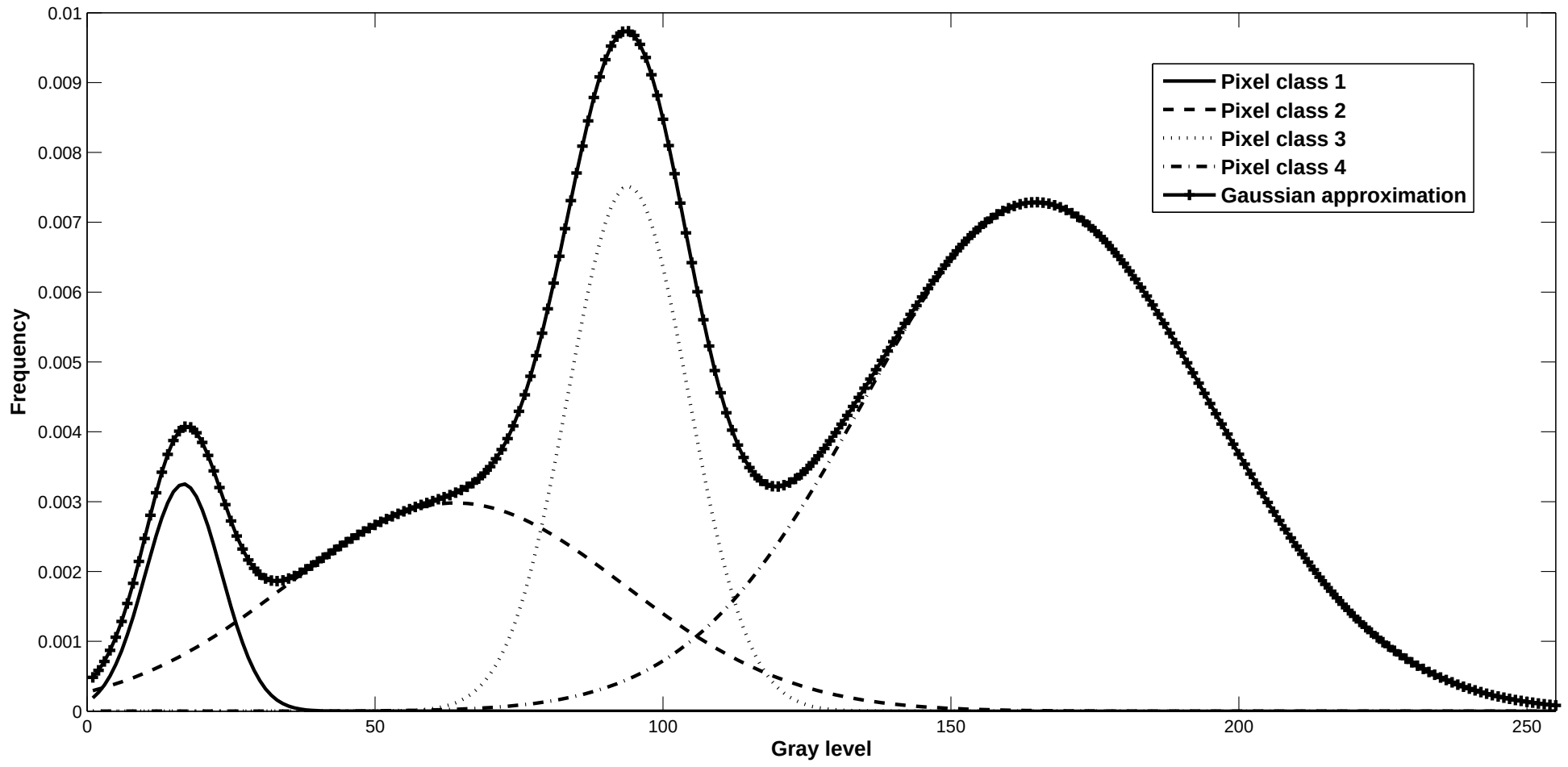

**Figure 4.** Gaussian functions obtained by LA.

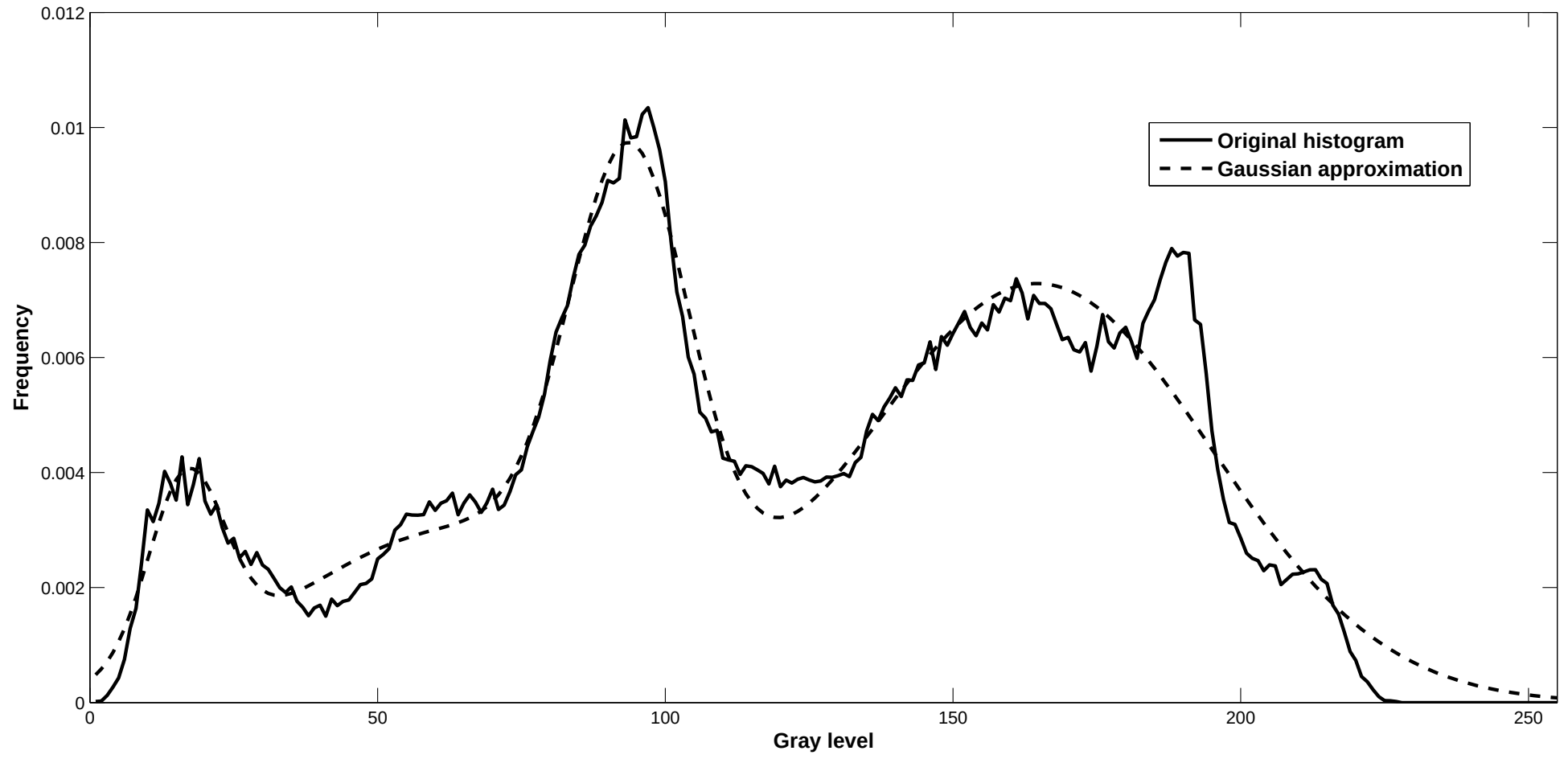

**Figure 5.** Comparison between the original histogram and its approach.

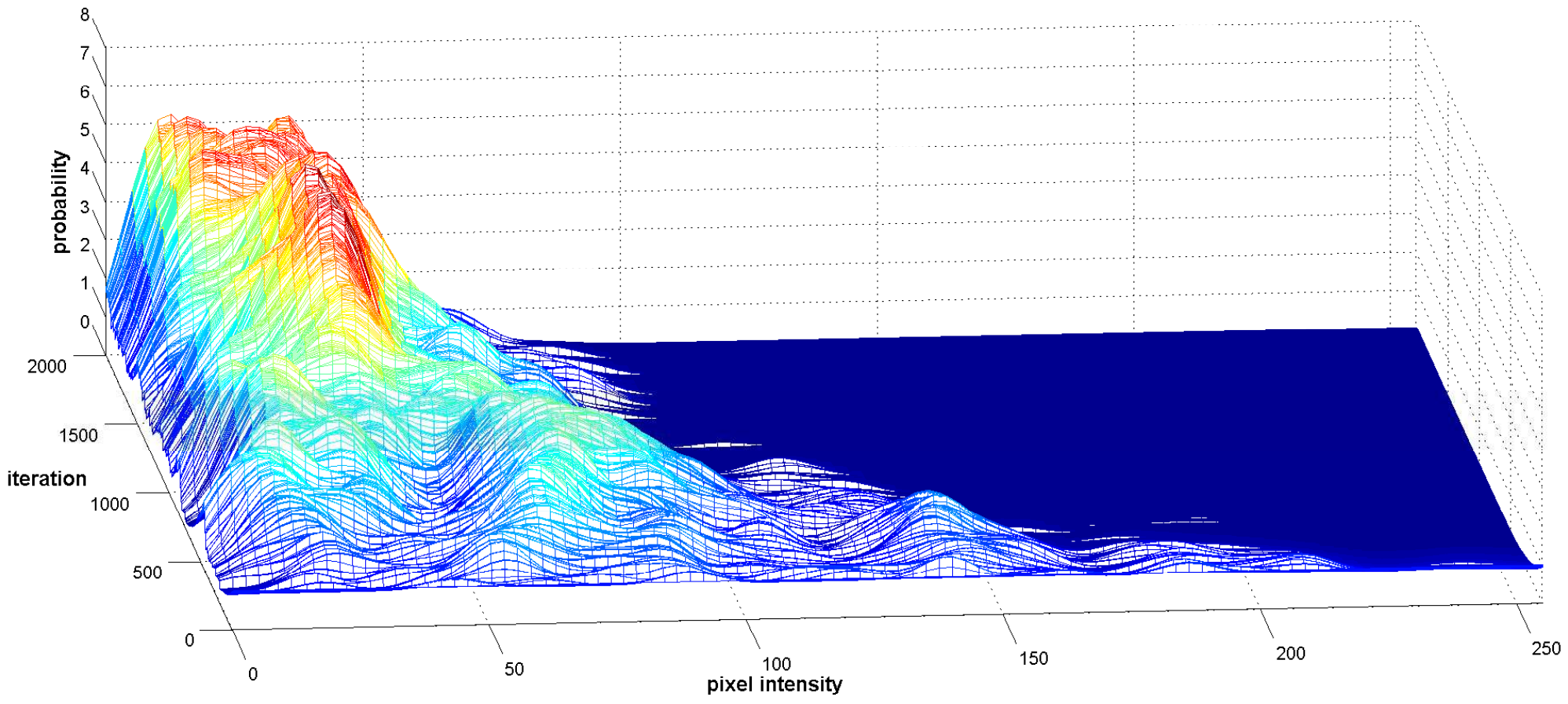

(a)

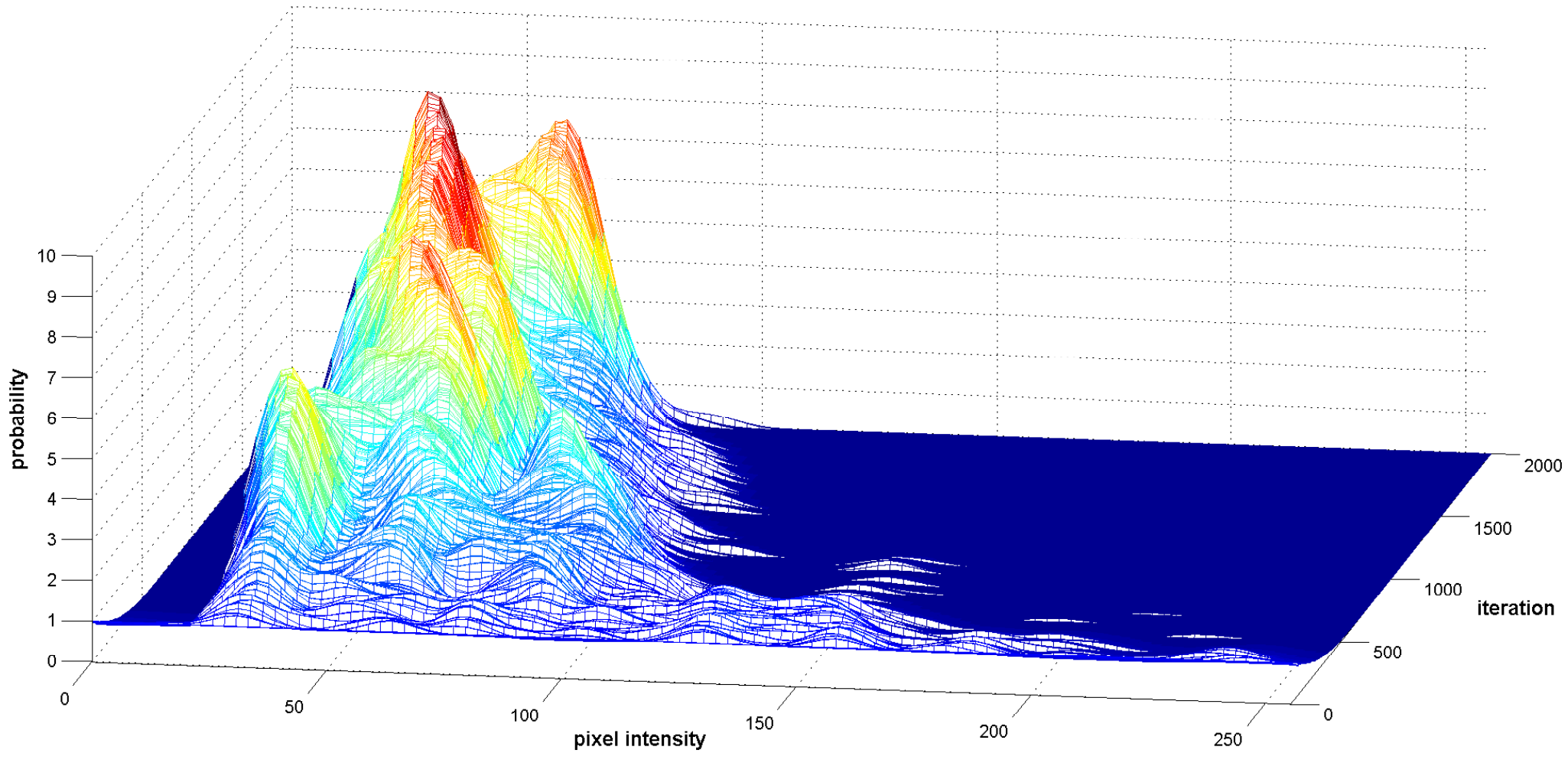

(b)

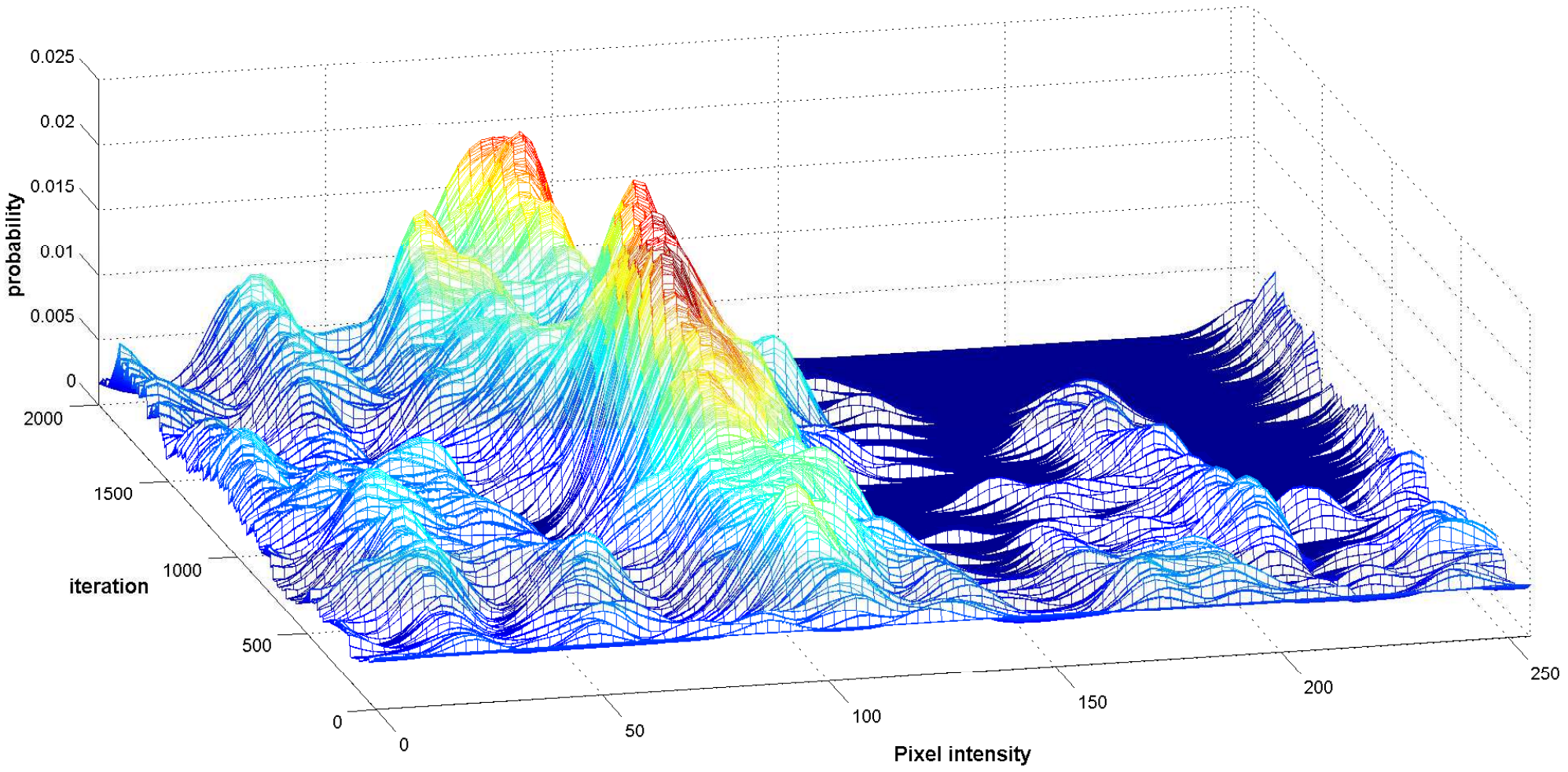

(c)

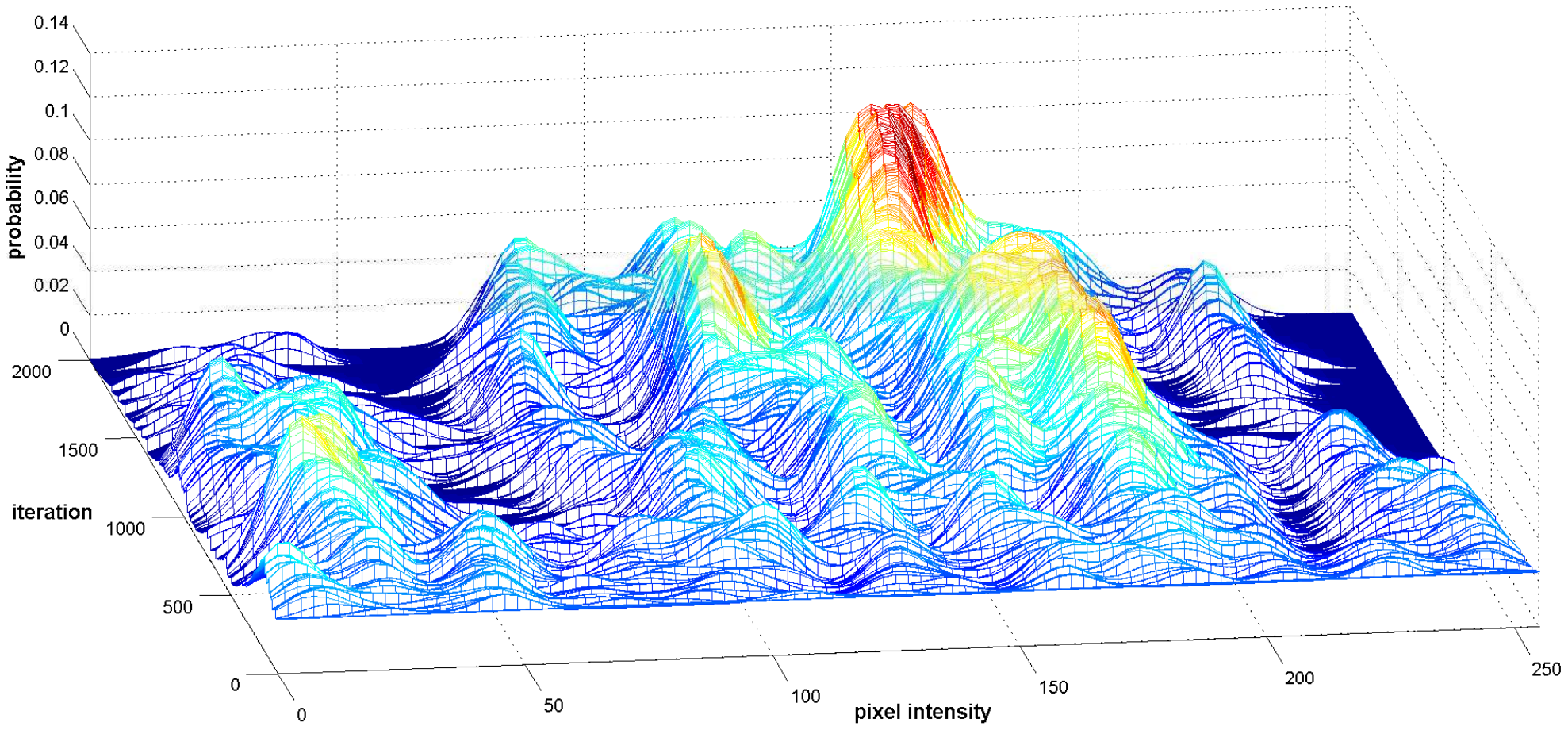

(d)

**Figure 6.** Evolution of the probability densities parameters and their expected values of the Gaussian functions (a) $f(k_{\mu}^{1},n)$, (b) $f(k_{\mu}^{2},n)$, (c) $f(k_{\mu}^{3},n)$ and (d) $f(k_{\mu}^{4},n)$.

From the Gaussian functions obtained by LA in Figure 4, the threshold values $T_i$ are calculated using well-known methods. Considering such values, the segmented image is shown in Figure 7.

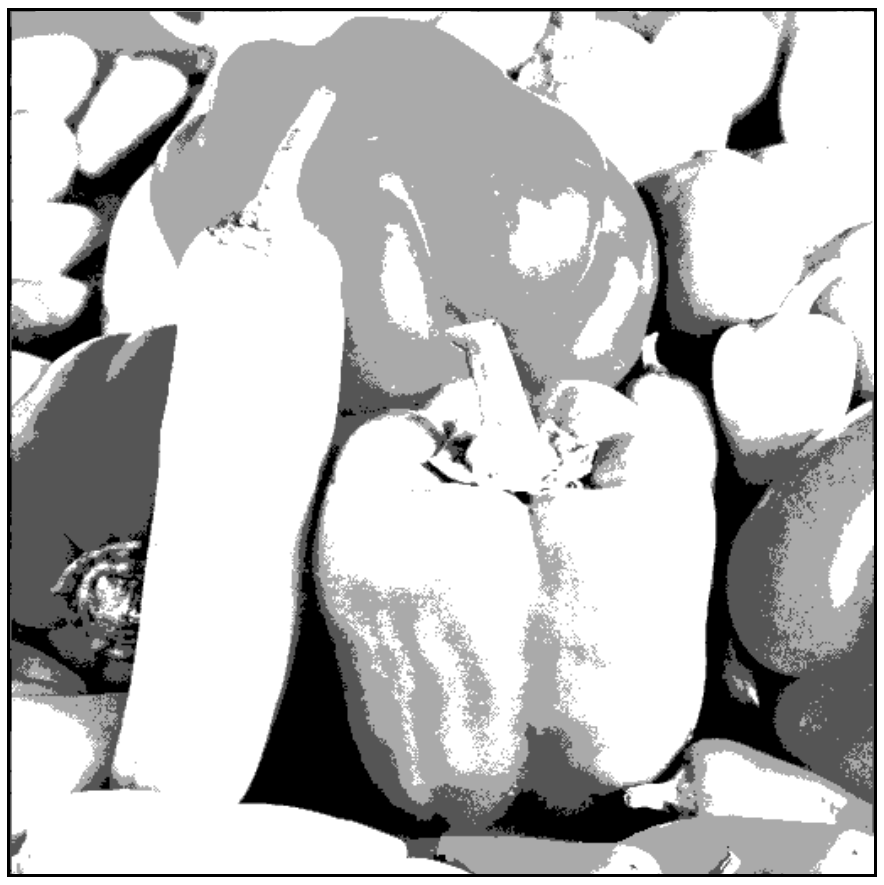

**Figure 7.** Image segmented in four classes by the LA method.

For the second experiment the image shown in Figure 8 is tested. The method aims to segment the image into four different classes using the LA approach. After executing the algorithm according to the parameters defined in Section 4, the resulting Gaussian functions approximating the histogram are shown in Figure 9a.

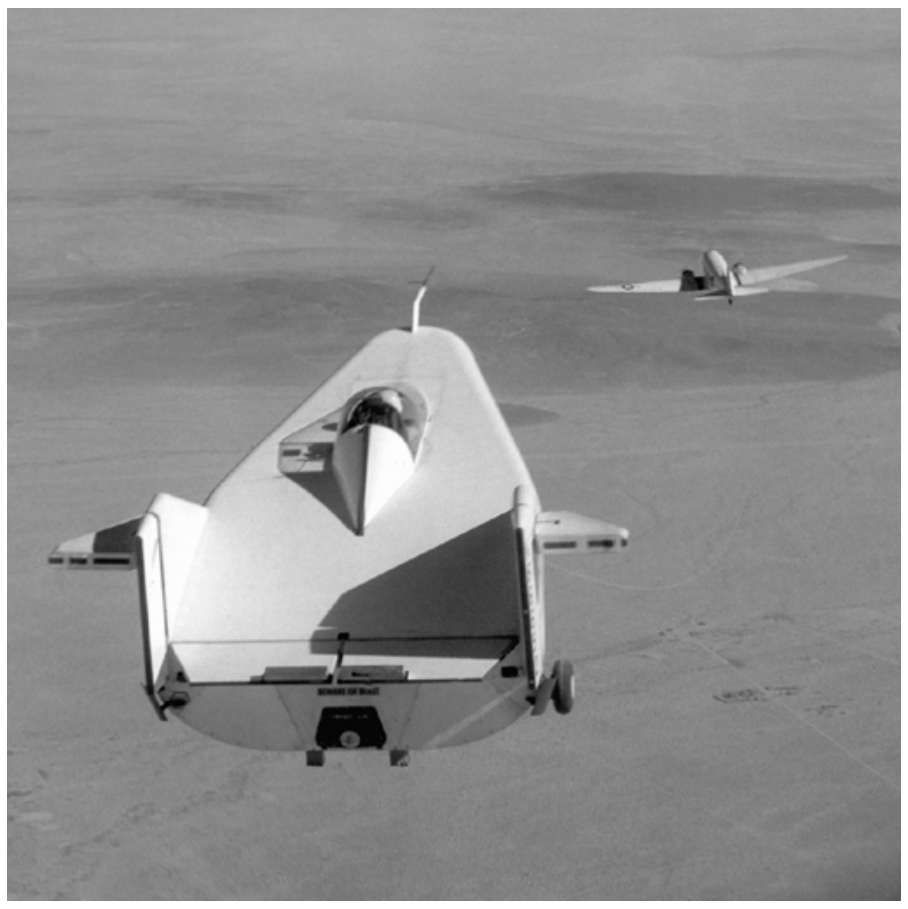

**Figure 8.** Original image used in the second experiment.

The comparison between the original image and its Gaussian approximation is shown in Figure 9b. It is clear that the algorithm approaches each of all the pixel concentrations distributed within the histogram but the first one, which is presented approximately around the intensity value seven. This effect shows that the algorithm discards the smallest pixel accumulation as it prefers to cover classes that contribute to generate smaller errors during optimization of the Equation 3. Such results can improve if five pixel classes were used instead.

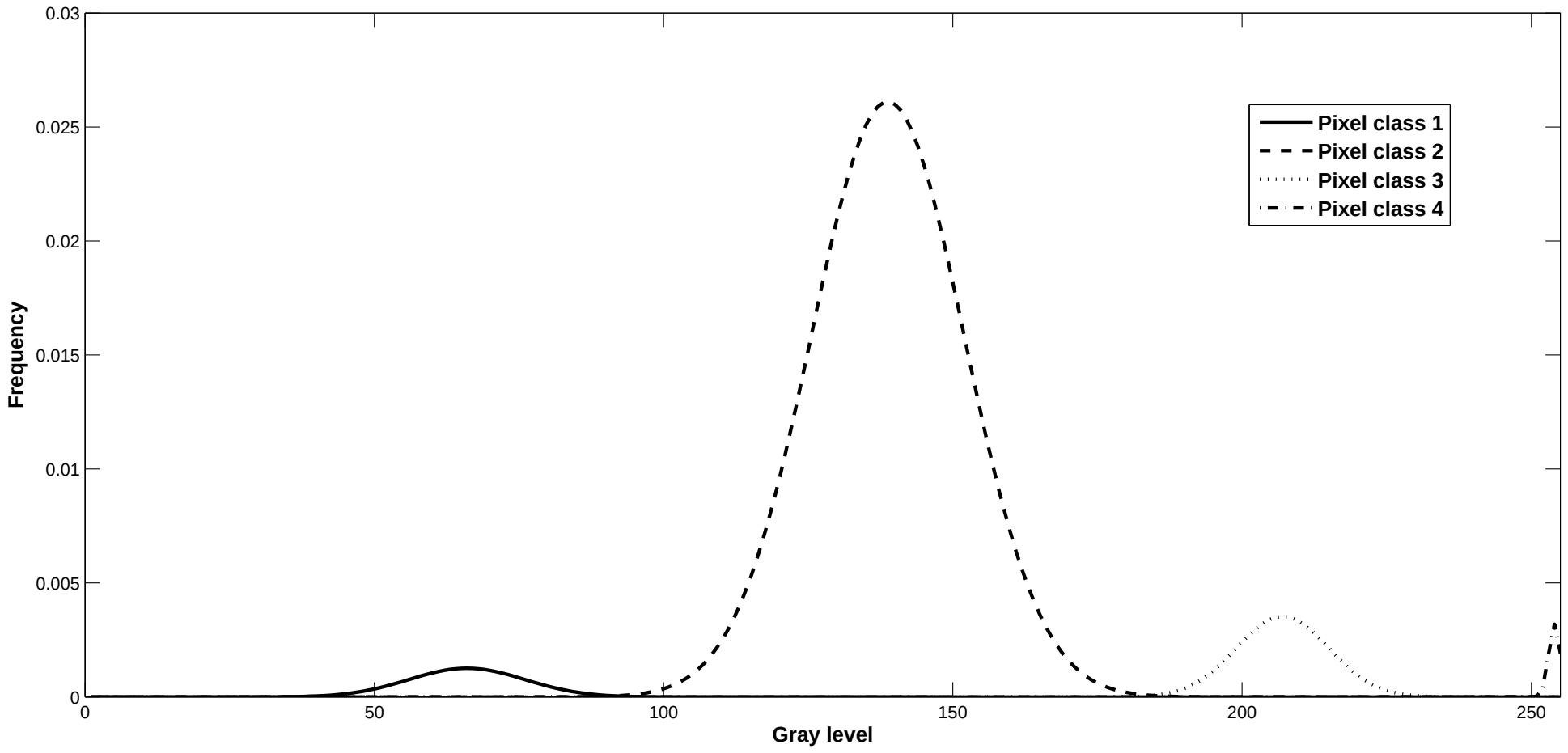

(a)

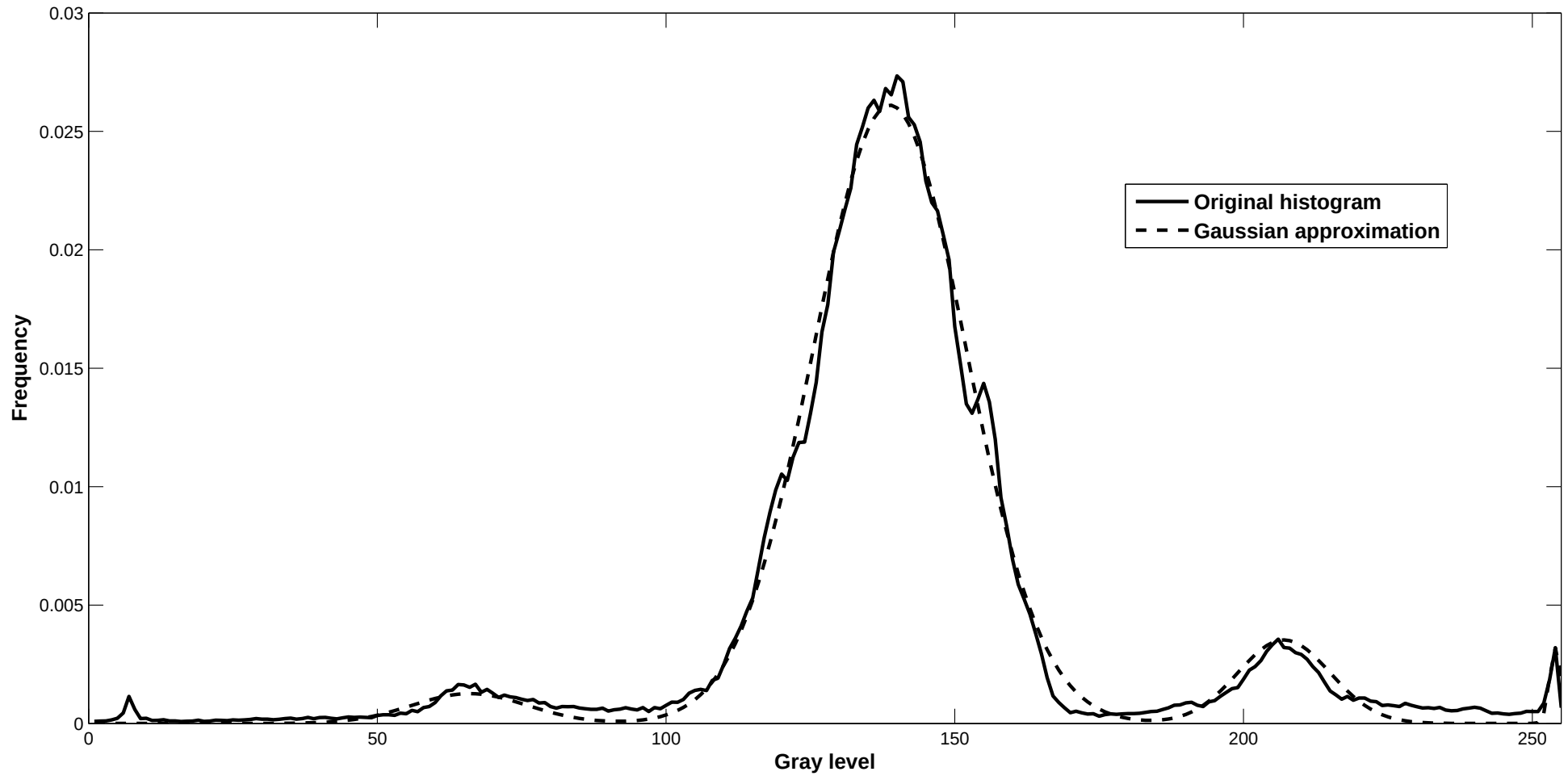

(b)

Figure 9. (a) Gaussian functions obtained by the LA algorithm and (b) its comparison to the original histogram.

From the Gaussian mixture obtained by the LA method (Figure 9a), the threshold values $T_i$ are calculated again using well-known methods. Figure 10 shows the segmented image after the detection task. Figure 11 shows the separation of each class after applying the LA algorithm.

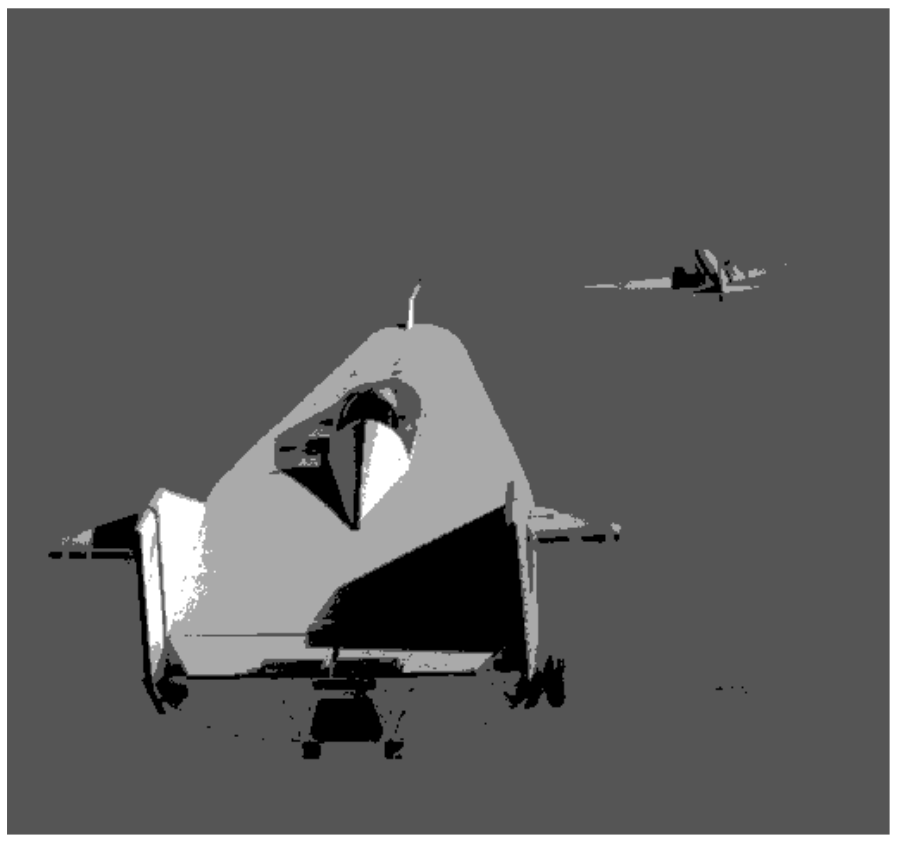

**Figure 10.** Segmentation obtained by LA.

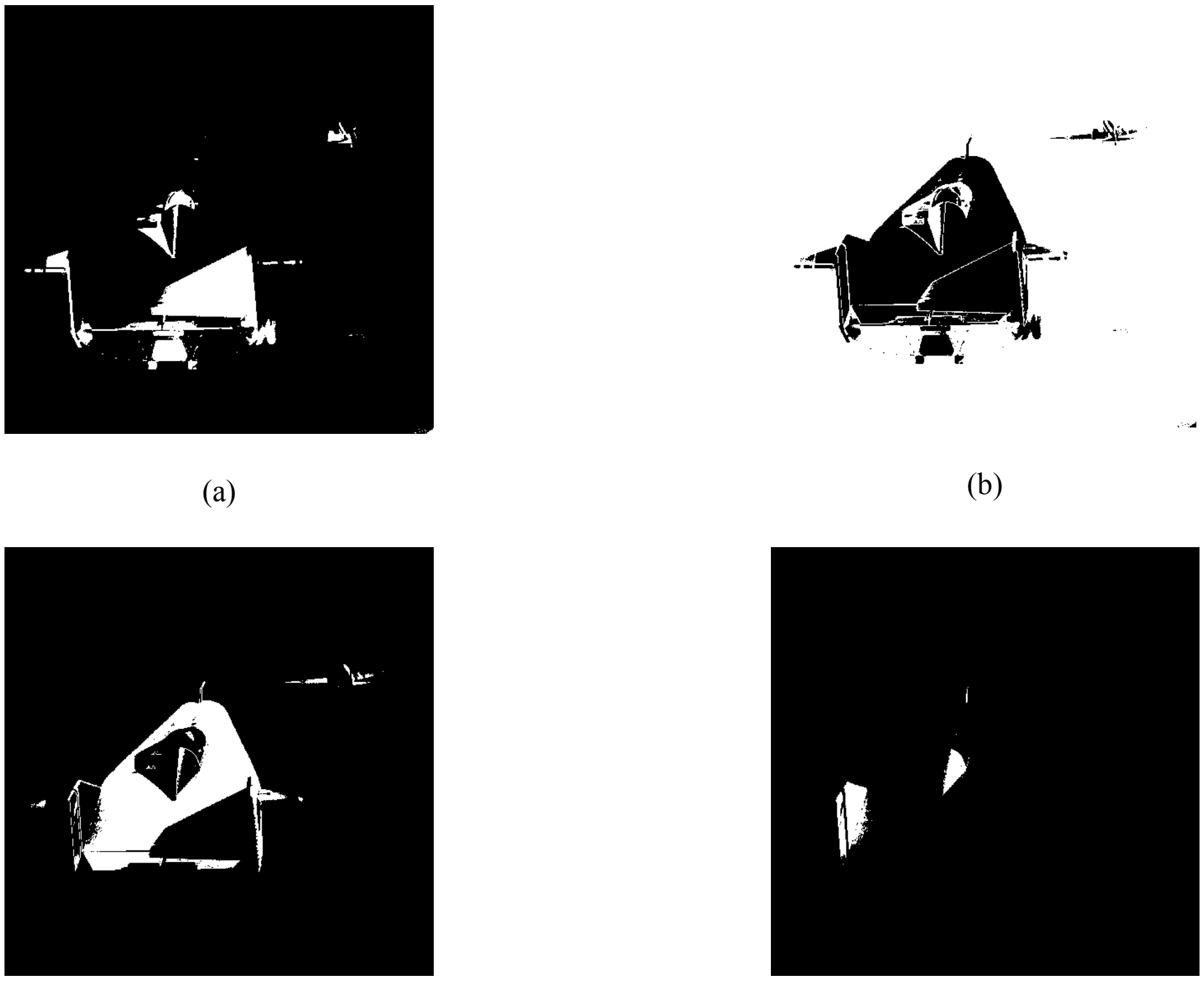

(a) (b) (c) (d)

**Figure 11**. Class separation as it is produced by the LA algorithm. (a) Pixel class 1, (b) Pixel class 2, (c) Pixel class 3, and (d) Pixel class 4.

*5.2 Comparing the LA algorithm vs. the EM and LM methods.*

This section discusses on the comparison between LA and other algorithms such as the EM algorithm and one Levenberg-Marquardt (LM) method. The discussion is focused on the following issues: first, sensitivity to the initial conditions; second, singularities and third, convergence and computational costs.

*a) Sensitivity to the initial conditions.* In this experiment, initial values for all methods are initialized in different values while the same histogram is considered for the approximation task. The final parameters representing the Gaussian mixture after convergence are reported. Figure 12a shows the image used in this comparison while Figure 12b pictures the histogram. All experiments are conducted several times in order to assure consistency. Only two different initial states with the highest variation are reported in Table 3. Likewise, Figure 13 shows the obtained segmented images considering the two initial conditions reported by Table 3. In the LA case, the algorithm does not require initialization as it works with random initial values; however in order to assure a valid comparison, the same initial values are considered for the EM, the LM and the LA method.

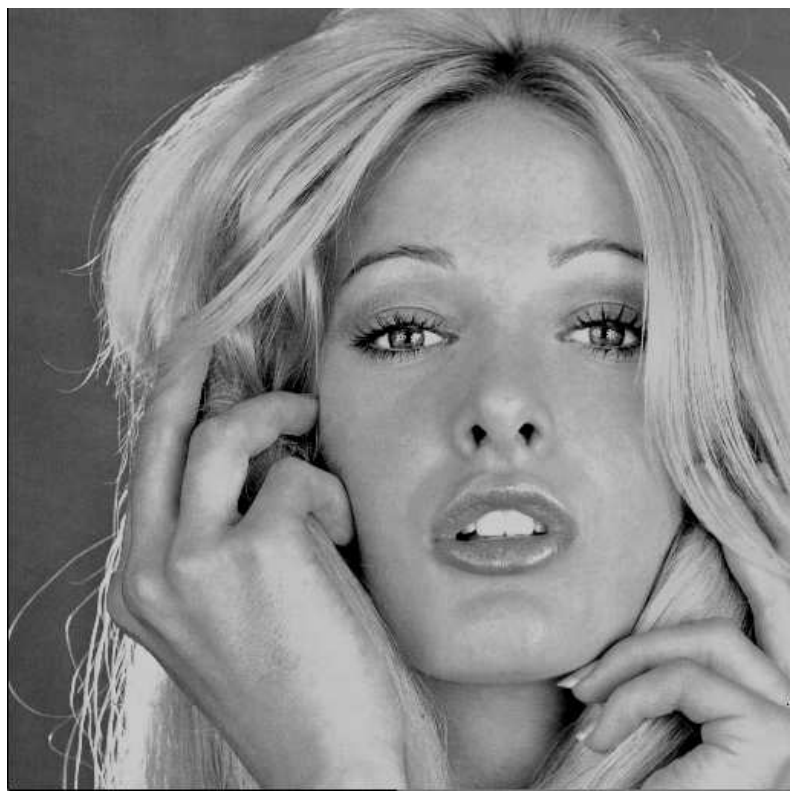

(a)

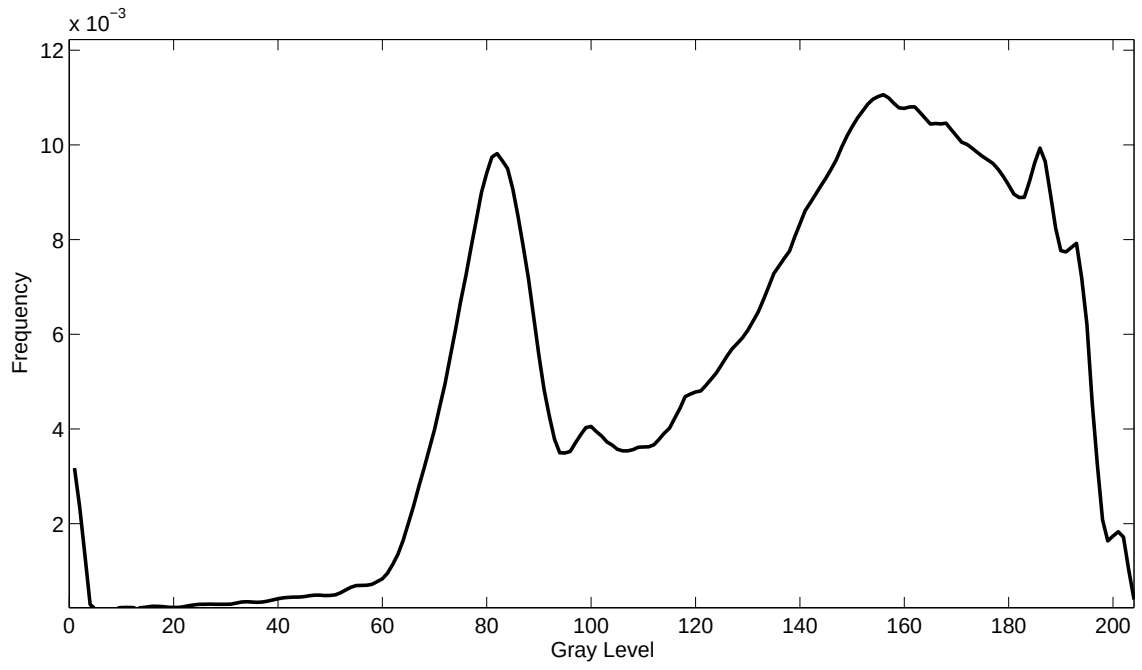

(b)

**Figure 12.** (a) Original image used for the comparison on initial conditions and (b) its corresponding histogram.

By analyzing the information in Table 3, the sensitivity of the EM algorithm to initial conditions becomes evident. Figure 13 shows a clear pixel misclassification in some sections of the image as a consequence of such sensitivity.

| Parameters | Initial condition 1 | EM | LM | LA | Initial condition 2 | EM | LM | LA |
|---|---|---|---|---|---|---|---|---|
| $k_\mu^1$ | 40.6 | 33.13 | 32.12 | 32.10 | 10 | 20.90 | 31.80 | 32.92 |
| $k_\mu^2$ | 81.2 | 81.02 | 82.05 | 82.01 | 100 | 82.78 | 80.85 | 82.12 |
| $k_\mu^3$ | 121.8 | 127.52 | 127 | 126.95 | 138 | 146.67 | 128 | 127.01 |
| $k_\mu^4$ | 162.4 | 167.58 | 166.80 | 166.72 | 200 | 180.72 | 165.90 | 166.62 |
| $k_\sigma^1$ | 15 | 25.90 | 25.50 | 25.51 | 10 | 18.52 | 20.10 | 25.11 |
| $k_\sigma^2$ | 15 | 9.78 | 9.70 | 9.66 | 5 | 12.52 | 9.81 | 9.68 |
| $k_\sigma^3$ | 15 | 17.72 | 17.05 | 17.10 | 8 | 20.5 | 15.15 | 17.12 |
| $k_\sigma^4$ | 15 | 17.03 | 17.52 | 17.55 | 22 | 10.09 | 18.00 | 17.15 |
| $k_P^1$ | 0.25 | 0.0313 | 0.0310 | 0.312 | 0.20 | 0.0225 | 0.0312 | 0.312 |
| $k_P^2$ | 0.25 | 0.2078 | 0.2081 | 0.2078 | 0.30 | 0.2446 | 0.2079 | 0.2088 |
| $k_P^3$ | 0.25 | 0.2508 | 0.2500 | 0.2510 | 0.20 | 0.5232 | 0.2502 | 0.2500 |
| $k_P^4$ | 0.25 | 0.5102 | 0.5110 | 0.5103 | 0.30 | 0.2098 | 0.5108 | 0.5103 |

**Table 3.** Comparison between the EM, the LM and the LA algorithm, considering two different initial conditions.

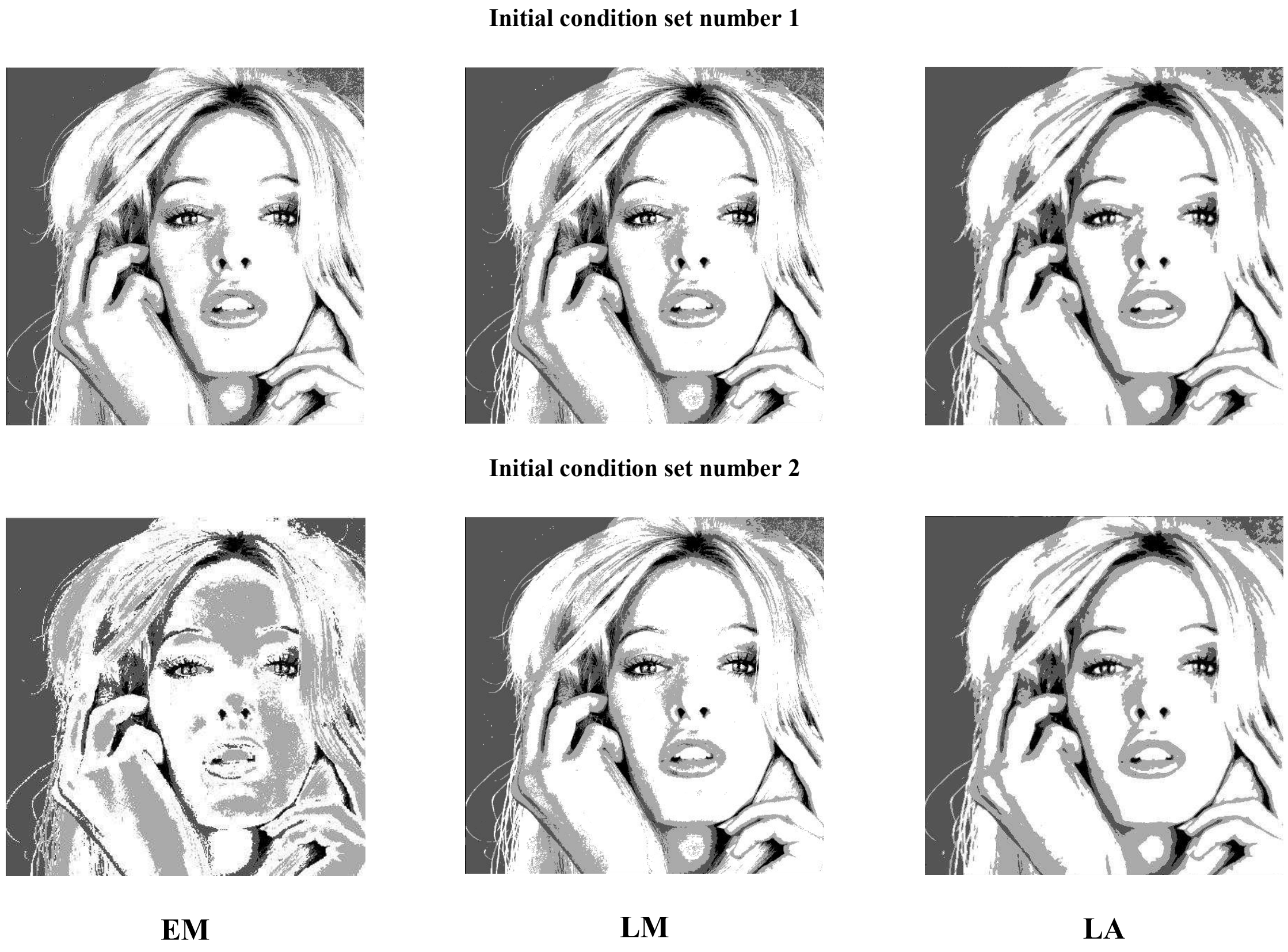

**Figure 13.** Segmented images after applying the EM, the LM and the LA algorithm with different initial conditions.

*b) Singularities.* The experiment aims to test the LA performance under certain circumstances on which it is well-reported in the literature [26,29] that the EM and the LM have underperformed. Two cases are relevant to such purpose. First, the Gaussian variance is small or near to zero, i.e. big objects are present in the image with a homogeneous intensity value [26]. Second, the LM algorithm exhibits a slow convergence when the Gaussians are overlapped [29, 33]. For both cases, the EM method never reaches convergence. The benchmark image and its histogram are shown in Figure 14

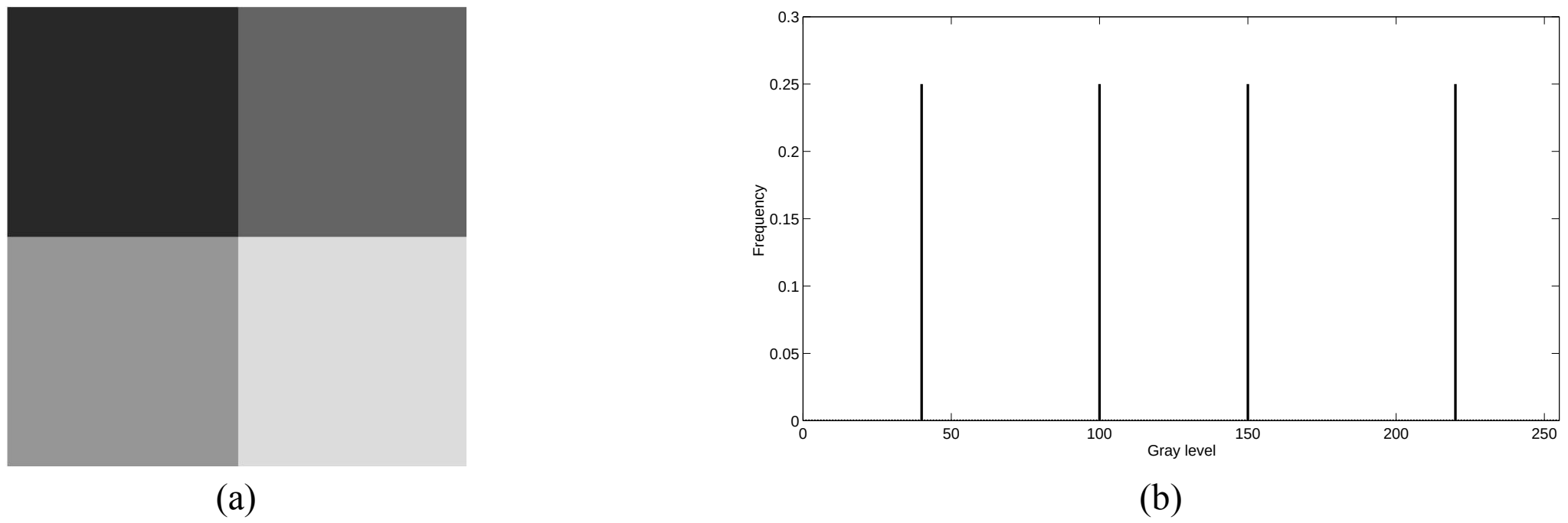

**Figure 14.** (a) Original image used by the singularity experiment, and (b) its histogram.

Case 1. The experiment shows the lack of convergence of the EM algorithm when a small or near to zero Gaussian variance is considered. The test consists on using all the algorithms to obtain the Gaussian mixture parameters that approximate the histogram shown in the Figure 14b. It is evident that only 4 classes are relevant. In order to assure consistency, the experiment is repeated over 100 times with different initial conditions. The results show that the EM method never converge to an acceptable value whatsoever. Table 4 shows the results for the LM and the LA algorithm as they are averaged over 100 experiments.

| Parameters | LM | LA |
|---|---|---|
| $\bar{k}_{\mu}^{1}$ | 42.6 | 40.1 |
| $\bar{k}_{\mu}^{2}$ | 98.3 | 99.89 |
| $\bar{k}_{\mu}^{3}$ | 153.7 | 150.05 |
| $\bar{k}_{\mu}^{4}$ | 220.1 | 220.01 |
| $\bar{k}_{\sigma}^{1}$ | 7 | 0.05 |
| $\bar{k}_{\sigma}^{2}$ | 12 | 0.07 |
| $\bar{k}_{\sigma}^{3}$ | 5 | 0.10 |
| $\bar{k}_{\sigma}^{4}$ | 0.3 | 0.03 |
| $\bar{k}_{P}^{1}$ | 0.20 | 0.0313 |
| $\bar{k}_{P}^{2}$ | 0.3 | 0.2078 |
| $\bar{k}_{P}^{3}$ | 0.25 | 0.2508 |
| $\bar{k}_{P}^{4}$ | 0.25 | 0.5102 |
| iterations | 997 | 1050 |

**Table 4.** Comparison between the LM and the LA algorithms using variances values close to zero.

By analyzing data in Table 4, it is clear that the LM and the LA algorithms are able to successfully segment the image shown in Figure 14a. The LM method converges a little faster than the LA algorithm. However, it shows a sub-optimal approximation to a local minimum (see Figure 15a).

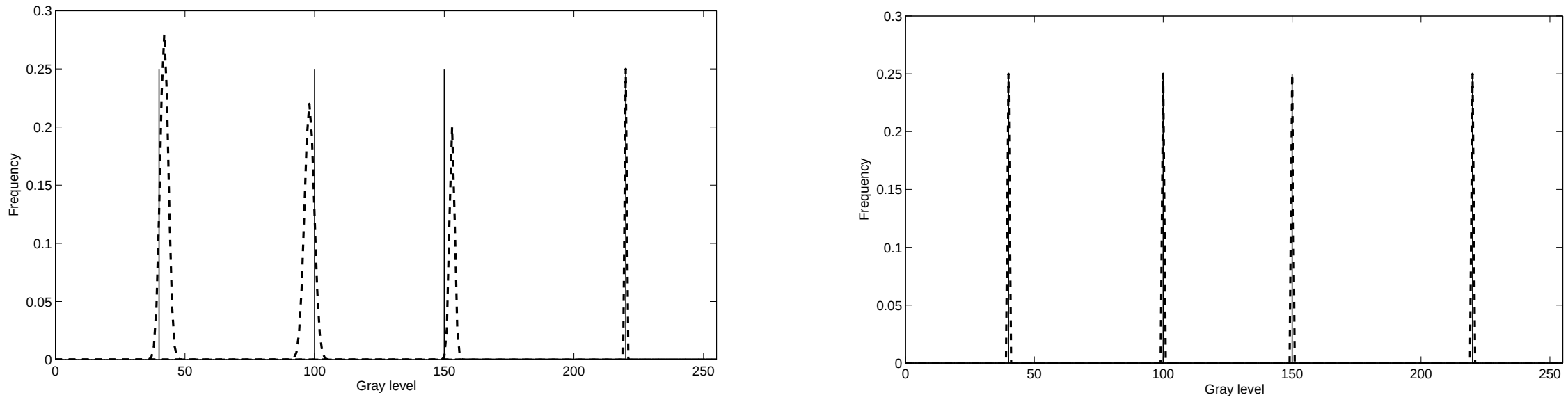

(a) (b)

**Figure 15.** Graphical view of approximations using near zero variances with: a) the LM algorithm and b) the LA method.

Case 2. This case analyzes the slow convergence of the LM method when the parameters of the Gaussian mixture are overlapped. For the experiment, the Gaussian's overlapping is caused by considering initial values falling on the same position. Although the results fully match with those in Case 1 (see Table 4), the differences on the required iterations are evident. For instance, the LA method requires nearly 1000 iterations while the LM method as much as 2300 iterations -averaging 100 experiments for both cases. The convergence speed in the LA method is clearly not affected by such singularity.

*c) Convergence and computational cost.* The experiment aims to measure the number of required steps and the computing time spent by the EM, the LM and the LA algorithm required to calculate the parameters of the Gaussian mixture in benchmark images (see Figure 16a-c). All experiments consider four classes. Table 5 shows the averaged measurements as they are obtained from 20 experiments. It is evident that the EM is the slowest to converge (iterations) and the LM shows the highest computational cost (time elapsed) because it requires complex Hessian approximations. On the other hand, the LA shows an acceptable compromise between its convergence time and its computational cost. Finally, Figure 16 below shows the segmented images as they are generated by each algorithm.

| Iterations / Time elapsed | (a) | (b) | (c) | (d) |
|---|---|---|---|---|
| **EM** | 1855 | 1833 | 1861 | 1870 |
| | 2.72s | 2.70s | 2.73s | 2.73s |
| **LM** | 985 | 988 | 945 | 958 |
| | 4.03s | 4.04s | 4.98s | 4.98s |
| **LA** | 970 | 991 | 951 | 951 |
| | 1.51s | 1.53s | 1.48s | 1.48s |

**Table 5.** Iterations and time requirements of the EM, the LM and the LA algorithm as they are applied to segment benchmark images (see Figure 16).

## 6. Conclusions

In this paper, an automatic image multi-threshold approach based on Learning Automata (LA) is proposed. The segmentation process is considered to be similar to an optimization problem. The algorithm approximates the 1-D histogram of a given image using a Gaussian mixture model whose parameters are calculated through the LA algorithm CARLA. Each Gaussian function approximating the histogram represents a pixel class and therefore one threshold point.

Experimental evidence shows that LA algorithm has an acceptable compromise between its convergence time and its computational cost when it is compared to the Expectation-Maximization (EM) method and the Levenberg-Marquardt (LM) algorithm. Additionally, the LA algorithm also exhibits a better performance under certain circumstances (singularities) on which it is well-reported in the literature [26,29] that the EM and the LM have underperformed. Two cases are reported: First, when Gaussian variance is small or near to zero (i.e. big objects are presented on the image with a homogeneous intensity value). Second, it is when the parameters of the Gaussian mixture are overlapped. Finally, the results have shown that the stochastic search accomplished by the LA method shows a consistent performance with no regard of the initial value and still showing a greater chance to reach the global minimum.

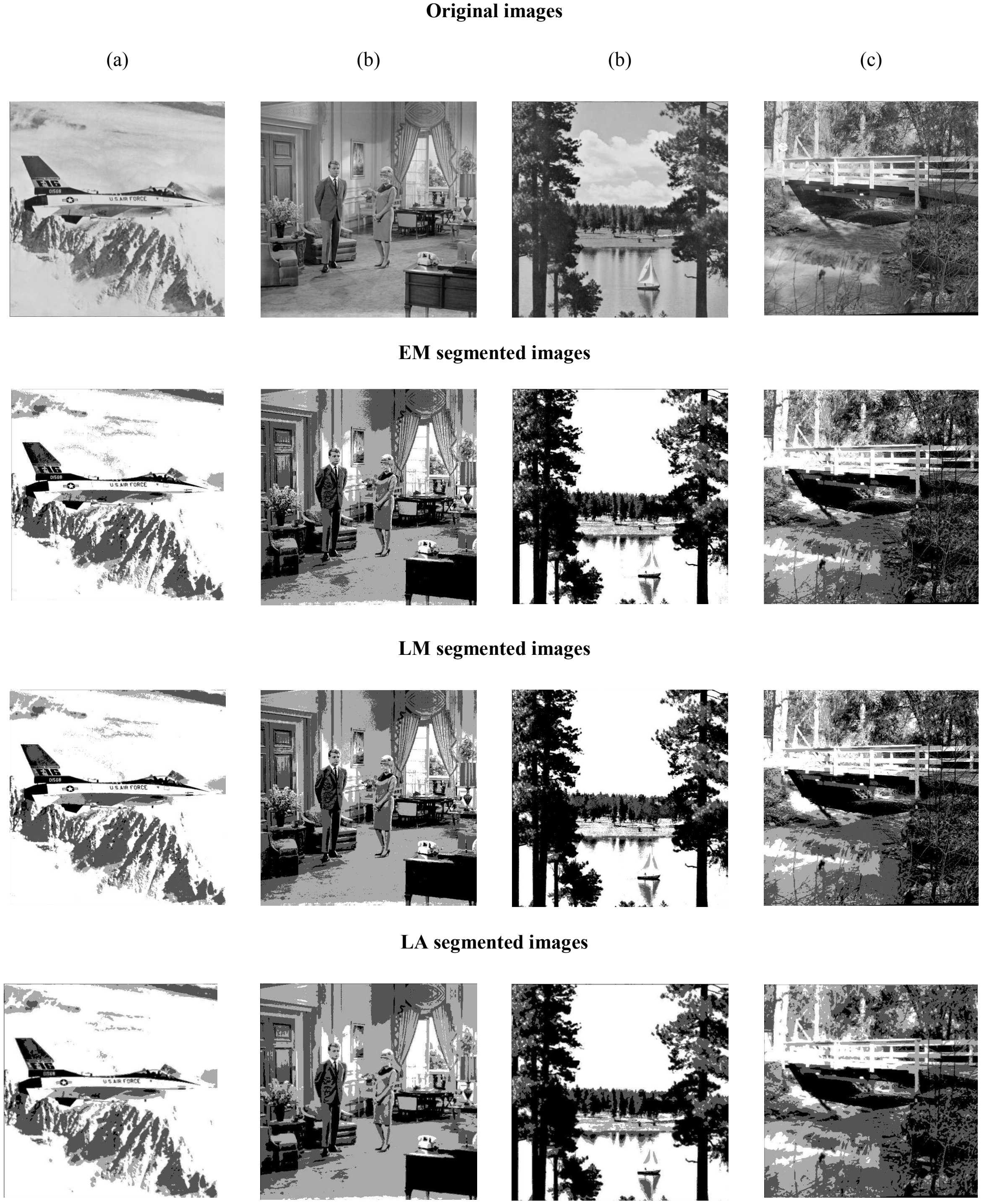

**Figure 16.** Original benchmark images a)-c), and segmented images obtained by the EM, the LM and the LA algorithms.